\documentclass[9pt]{article}
\usepackage[margin=1.25in]{geometry}
\usepackage{graphicx} 
\usepackage[square,sort,comma,numbers]{natbib}
\usepackage{booktabs}
\usepackage{multirow}
\usepackage{enumitem}
\usepackage{authblk}
\usepackage{amssymb}
\usepackage{hyperref}
\usepackage{tipa}
\usepackage{array}

\usepackage{tikz}
\usepackage{amsmath, amssymb}
\usetikzlibrary{shapes.geometric, arrows.meta, positioning, calc, backgrounds, fit}
\usepackage{xcolor}

\usepackage{tabularx}

\definecolor{processblue}{RGB}{66, 133, 244}
\definecolor{injusticered}{RGB}{219, 68, 55}
\definecolor{injusticeorange}{RGB}{244, 160, 0}
\definecolor{gapyellow}{RGB}{255, 235, 180}
\definecolor{lightgray}{RGB}{245, 245, 245}

\begin{document}

\title{Beyond Single Ground Truth: Reference Monism as Epistemic Injustice in ASR Evaluation}
\author[a,1,*]{Anna Seo Gyeong Choi}
\author[b]{Maria Teleki} 
\author[b]{James Caverlee}
\author[c,2]{Miguel del Rio}
\author[d,2]{Corey Miller}
\author[e,2]{Hoon Choi}

\affil[a]{Department of Information Science, Cornell University}
\affil[b]{Department of Computer Science, Texas A\&M University}
\affil[c]{Rev AI}
\affil[d]{Tundra Technical Solutions}
\affil[e]{Division of Liberal Studies, Kangwon National University}

\affil[1]{To whom correspondence should be addressed. E-mail: sc2359@cornell.edu}
\affil[2]{These authors contributed equally.}
\affil[*]{Work conducted in part during an internship at Rev AI.}

\date{}
\maketitle
\begin{abstract}
Automatic speech recognition (ASR) evaluation compares system output to ground truth transcripts, with Word Error Rate (WER) quantifying the distance between them. But ground truth transcripts are not discovered -- they are produced by human annotators following conventions that encode normative assumptions about which speech features matter. Different conventions (verbatim, non-verbatim, legal) produce different transcripts of identical speech and judge the same ASR output differently.
This paper argues that \textbf{reference monism} -- enforcing a single transcription convention as ground truth -- commits \textbf{epistemic injustice}. Speakers with aphasia, whose speech includes clinically meaningful disfluencies, are systematically disadvantaged when evaluated against ``clean'' references that treat those disfluencies as errors. The harm is not merely differential performance, but that evaluative infrastructure lacks interpretive resources to recognize their contributions as legitimate. We develop a philosophical framework introducing the \textit{hermeneutical gap}, formalize \textit{Epistemic Injustice Distance} (EID) to measure reference monism's cost, and demonstrate empirically using AphasiaBank that WER varies depending on which convention defines ground truth. We propose \textbf{WER-Range}: reporting performance across legitimate conventions rather than assuming a single correct answer.

\end{abstract}


\section{Introduction}

How should we evaluate Automatic Speech Recognition (ASR) systems?  ASR systems mediate access to essential services -- voice assistants transcribe queries, clinical documentation systems record patient encounters, accessibility tools caption lectures and meetings. As these systems become infrastructure, their evaluation practices determine whose speech is deemed ``recognizable''  as-is and whose speech is treated as a problem to be solved. The standard methodology appears straightforward: a system produces a hypothesis; evaluators compare it to a ground truth transcript; Word Error Rate (WER) quantifies the distance between them. Lower is better. This framework underlies benchmark construction, fairness audits, and deployment decisions \cite{eriksson2025trustaibenchmarksinterdisciplinary, koenecke2020racial, dheram2022toward}.\footnote{WER dominates contemporary ASR evaluation, evidenced by its status as the sole quality metric on the Open ASR Leaderboard: \url{https://huggingface.co/spaces/hf-audio/open_asr_leaderboard}}
We focus specifically on \textit{evaluation practices}: how researchers, auditors, and developers assess system quality. This is distinct from questions about what transcripts ASR systems should produce for end users -- a system may legitimately output clean text for accessibility applications while being evaluated against multiple reference conventions to ensure fair assessment across speaker populations.

But ground truth transcripts are not discovered but constructed -- a Kantian insight about the nature of knowledge \cite{kant1909critique, kant1999critique}: we do not passively apprehend speech events as objective facts, but actively constitute them through interpretive frameworks. Ground truth transcripts are produced by human annotators following conventions that encode normative assumptions about which features of speech matter. As Bucholtz \cite{bucholtz2000politics} argues, transcription is inherently political: choices about what to preserve, normalize, or exclude reflect and reproduce power relations between speakers and institutions. A verbatim transcript preserves fillers, false starts, and repairs. A non-verbatim transcript removes these, producing ``clean'' text. A legal transcript preserves hedges relevant for evidentiary purposes.\footnote{These convention types follow typical transcription standards. See Rev AI transcription guidelines \url{https://www.rev.com/resources/verbatim-transcription} and legal transcription standards \url{https://www.legallanguage.com/legal-articles/the-4-rules-of-legal-transcription/}.} Each convention serves legitimate purposes, produces a different transcript of the same utterance, and judges the same ASR output differently.

\begin{figure*}[t]
    \centering
    \includegraphics[width=\textwidth]{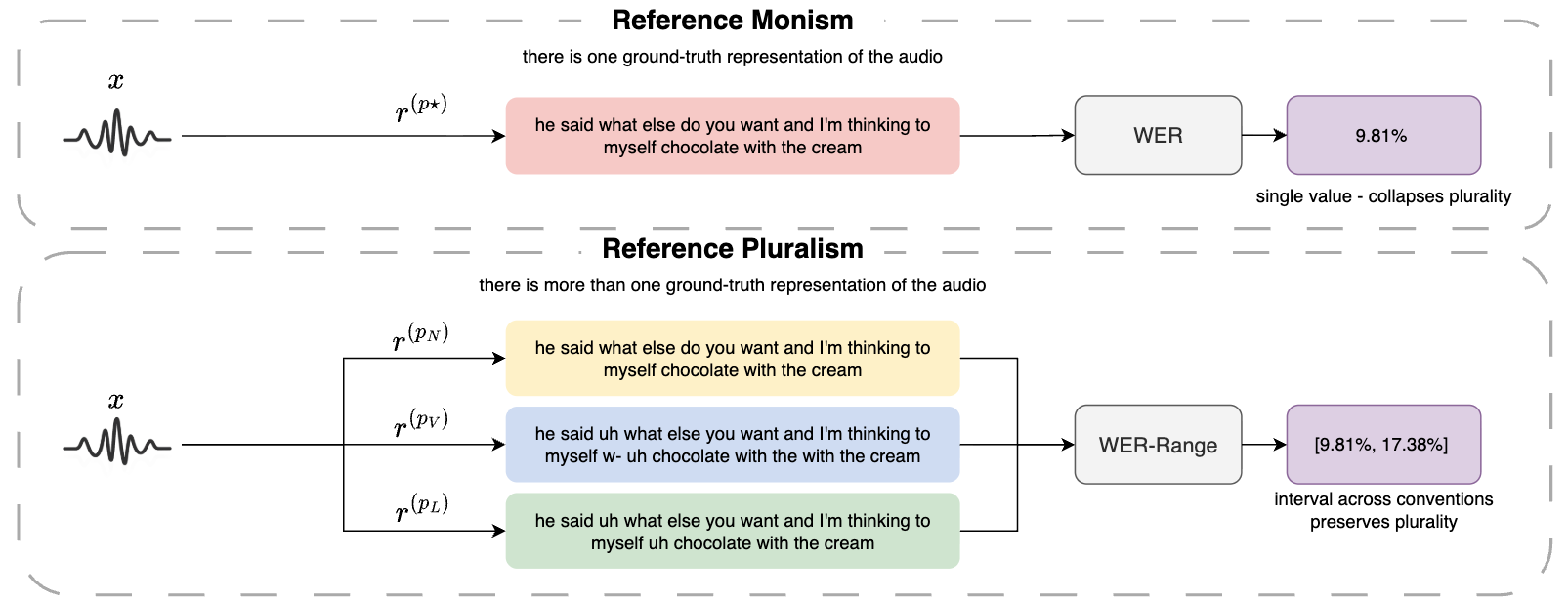}
    \caption{Reference monism versus pluralism, and their metric consequences. \textit{Top}: under reference monism, a single enforced convention $r(p^\star)$ produces one WER score (9.81\%), naturalizing one interpretive choice as ground truth and collapsing legitimate plurality into a single number. \textit{Bottom}: under reference pluralism, the same utterance yields three transcripts under distinct conventions -- \textit{non-verbatim} $r(p_N)$ removes all disfluencies; \textit{verbatim} $r(p_V)$ preserves fillers, fragments, and repairs exactly as spoken; \textit{legal} $r(p_L)$ retains hesitation markers but normalizes repairs -- which together feed into WER-Range, producing an interval [9.81\%, 17.38\%] that makes the convention-dependence of ``accuracy'' visible. The same ASR hypothesis $h(x)$ and audio $x$ underlie both computations; only the evaluative infrastructure differs.}
    \label{fig:monism-pluralism}
\end{figure*}


This paper argues that \textit{reference monism} -- the enforcement of a single transcription convention as ground truth -- commits epistemic injustice \cite{fricker2007epistemic}. Speakers with aphasia, whose speech is characterized by clinically meaningful disfluencies, are a clear case: they are penalized when evaluated against ``clean'' references that treat those disfluencies as errors. The harm is not merely that systems perform worse on these populations, but that the evaluative infrastructure itself lacks resources to recognize their contributions as legitimate.

Prior work has documented substantial ASR performance disparities across racial groups \cite{koenecke2020racial}, dialects \cite{wassink2022uneven}, age groups \cite{vipperla2010ageing}, and clinical populations \cite{fraser2013automatic, zhao2025quantification, mei2025addressing}. Philosophical analysis of these disparities has identified ASR evaluation as a site of epistemic harm \cite{choi2025fairness}, yet these studies measure disparity against a fixed ground truth, attributing gaps to model limitations addressable through improved training data or architecture. Our contribution is orthogonal: we show that the choice of ground truth itself structures measured disparities, and that convention choice can be a source of injustice independent of model performance. Separately, a growing literature treats annotator disagreement as signal rather than noise \cite{plank2022problem, aroyo2015truth}, asking how to aggregate diverse judgments or preserve disagreement information. Where that work examines variation \textit{within} a convention (annotators following the same guidelines may still disagree), we examine variation \textit{across} conventions that legitimately produce systematically different labels. These perspectives are complementary: plural ground truth adds a dimension -- an interpretive framework -- that disagreement-aware approaches do not yet address.

Recent empirical work provides direct precedent for our theoretical claims. McNamara et al.\cite{mcnamara2024style} demonstrated that identical ASR output scores dramatically differently under verbatim versus non-verbatim references, showing WER variation for the same system-utterance pair; they note that machine translation adopted multi-reference evaluation decades ago, yet ASR has resisted this pluralism. Heuser et al. \cite{heuser2024quantification} showed that transcription style choices -- not acoustic modeling -- drive substantial measured disparities for African American English (AAE) speakers, with human transcribers' convention choices accounting for more variation than ASR system differences. These findings motivate the present work: we provide the philosophical framework explaining \textit{why} single-reference evaluation constitutes epistemic injustice and formalize \textbf{how} to measure its cost.

We develop this argument in three stages: (i) \textbf{a philosophical framework} introducing the \textit{hermeneutical gap} between speakers' contributions and conventions' interpretive resources; (ii) \textbf{a formalization} defining Epistemic Injustice Distance (EID) and $\Delta$EID to measure the cost of reference monism; and (iii) \textbf{empirical demonstration} using AphasiaBank \cite{aphasiabank}, showing that WER varies by nearly a factor of two depending on which convention defines ground truth. Our practical recommendation is \textbf{\textit{WER-Range}, a reporting practice} that reports performance across legitimate conventions rather than collapsing plurality into a single number. 
Figure~\ref{fig:monism-pluralism} illustrates the full argument: the same utterance yields three transcripts under distinct conventions, and the choice of which convention to enforce determines whether evaluation reports a single WER score or a WER-Range interval that makes that choice visible.

\section{Philosophical Groundwork}\label{sec:philosophy}

While ASR fairness audits have proliferated \cite{pradhan2020use, koenecke2020racial, lea2023user, zolnoori2024decoding, zhao2025quantification, mei2025addressing}, and benchmark criticism has identified validity concerns in AI evaluation generally \cite{eriksson2025trustaibenchmarksinterdisciplinary, wang2024benchmark, prandi2025bench2coptrustbenchmarkingeu, reuel2024betterbenchassessingaibenchmarks, aksenova2021might, likhomanenko2021rethinking}, no prior work has examined how the \textit{interpretive framework defining ground truth} structures measure disparities.
We draw on three philosophical traditions to argue that ASR evaluation commits a distinctive form of harm: \textit{epistemic injustice}. This section introduces the core concepts; fuller elaboration appears in Appendix~\ref{app:philosophy}.

\subsection{Epistemic Injustice}\label{sec:epistemic-injustice}

Fricker \cite{fricker2007epistemic} identifies harms done to individuals \textit{in their capacity as knowers} -- what she calls \textbf{epistemic injustice}. Unlike material or dignitary harms, epistemic harms damage one's ability to participate in knowing and communicating knowledge. Fricker distinguishes two forms:

\textbf{Testimonial injustice} occurs when a speaker receives a \textit{credibility deficit} -- their testimony is judged less believable than warranted -- due to prejudice rather than deficiency in the testimony itself. The wrong is being disbelieved \textit{because of who you are} rather than \textit{what you said}.

\textbf{Hermeneutical injustice} occurs when someone lacks the interpretive resources -- concepts, vocabulary, frameworks -- needed to make sense of their own experience, due to marginalization from collective meaning-making processes. Fricker's paradigm case is sexual harassment before the 1970s: women lacked shared vocabulary to make the experience socially intelligible, not because nothing happened, but because collective hermeneutical resources lacked adequate framework for it.
Dotson \cite{dotson2011tracking} distinguishes testimonial \textit{quieting} (external rejection of testimony) from testimonial \textit{smothering} (preemptive self-censorship anticipating low credibility). Harrington \cite{harrington2022s} document both in ASR: Black older adults explicitly described consciously modifying their speech as ``code-switching'' to be understood by voice assistants – testimonial smothering operating before systems can reject their natural speech.

Crucially, a gap in \textit{collective} resources does not mean no one has adequate tools. Goetze \cite{goetze2018hermeneutical} identifies \textbf{hermeneutical dissent}: cases where marginalized groups \textit{have} developed interpretive tools despite exclusion from dominant meaning-making. When such tools exist within a community but have not spread to other groups, Goetze terms this \textbf{hermeneutical ghettoization} -- community members understand their own experiences but cannot communicate them to outsiders who lack the interpretive resources. As we argue below, this precisely describes clinical speech communities and speakers of non-standard dialects vis-\`a-vis ASR systems.

\subsection{Structural Injustice and Willful Ignorance}

Anderson \cite{anderson2012epistemic} argues that individual epistemic virtue -- cultivating sensitivity to one's prejudices -- is insufficient when epistemic injustice is embedded in social institutions. Just as individual charity cannot remedy structural poverty, individual open-mindedness cannot remedy institutionalized hermeneutical gaps. 
This institutional perspective proves essential for ASR: hermeneutical gaps are embedded in transcription conventions developed without marginalized input \cite{koenecke2020racial}, benchmark datasets with demographic biases, evaluation metrics presupposing single correct transcriptions, and norms rewarding standard-benchmark performance \cite{eriksson2025can,reuel2024betterbench}. Notably, initiatives like Mozilla Common Voice \cite{ardila2020common} explicitly recognize these limitations, foregrounding community participation and documentation rather than treating conventions as neutral.

Pohlhaus \cite{pohlhaus2012relational} introduces \textbf{willful hermeneutical ignorance}: when dominant groups actively resist acquiring interpretive resources that marginalized communities have developed. Unlike simple hermeneutical injustice (gaps exist because no one developed adequate concepts), willful ignorance involves \textit{refusal} to adopt available tools. Clinical speech transcription conventions exist; sociolinguistic descriptions of AAE phonology are well-documented; disability communities have articulated communication norms for diverse speakers. Recent work by Rev AI demonstrates that commercial providers can incorporate diverse transcription conventions when motivated \cite{heuser2024quantification}, yet such efforts remain exceptional rather than standard practice. The continued absence of these resources from mainstream ASR evaluation is not mere oversight but structural refusal to expand the interpretive frameworks that determine what counts as accurate transcription.

\subsection{The Impossibility of Context-Free Ground Truth}

Dreyfus \cite{dreyfus1992computers} and Suchman \cite{suchman2006human} argue that human expertise operates through tacit, contextual judgment that cannot be captured in explicit rules.
Expert transcribers do not apply algorithms; they exercise holistic pattern recognition shaped by purpose, context, and background. \citet{love2021specifying} demonstrated this: eight trained forensic transcribers produced substantially different transcripts of identical audio -- not ``errors'' but different legitimate interpretive stances under acoustic uncertainty.
The assumption that transcription can be evaluated against a single, purpose-independent ground truth ignores this fundamental context-dependence. When trained transcribers disagree, they are not making ``errors'' relative to an objective standard; they are exercising different legitimate interpretive stances. This claim concerns disagreements arising from different transcription policies, not expertise asymmetries where one transcriber has domain knowledge the other lacks -- such as familiarity with dialectal features or regional terminology. Aggregating their judgments yields consensus, not objectivity.

Gadamer \cite{gadamer2013truth} argued that all understanding operates through \textit{prejudices} (Vorurteile) -- pre-judgments constituting the interpreter's ``horizon.'' These are not obstacles but enabling conditions: we make sense of something new by relating it to what we already understand. Applied to transcription, every ground truth embeds prejudices about what speech ``really says'': training data reflecting assumptions about ``standard'' speech, conventions treating disfluencies as deviations rather than legitimate acts, and metrics presupposing a single correct answer. These prejudices are not necessarily illegitimate -- conventions serve genuine purposes -- but they become sources of injustice when naturalized as objective standards rather than recognized as contestable choices.


\subsection{The Hermeneutical Gap}

Drawing these traditions together, we introduce the \textbf{hermeneutical gap}: the distance between a speaker's communicative contribution and the interpretive resources available in transcription conventions to render that contribution intelligible. For speakers of prestige dialects whose patterns align with conventions developed from and for their communities, the gap is minimal. For speakers of marginalized varieties -- marked by race, disability, class, or region -- the gap widens, and meaning is lost, distorted, or erased.

The hermeneutical gap is not a property of speakers but of \textit{evaluative frameworks}. It measures not how ``clearly'' someone speaks but how well the interpretive infrastructure accommodates their communicative practices. A speaker with aphasia communicates meaningfully within clinical contexts where verbatim conventions preserve disfluencies as diagnostically significant -- both the speaker's actual production and its transcription representation are valued for what they reveal about language processing. The same speaker appears to communicate poorly when evaluated against ``clean'' standards treating disfluencies as errors. The gap lies not in the speaker but in the mismatch between their production patterns and the interpretive frameworks (whether computational or conventional) used to render those patterns as text.

This concept synthesizes our philosophical resources: the gap constitutes hermeneutical injustice (Fricker) when it results from marginalization; it reflects ghettoization (Goetze) when marginalized communities have developed conventions that dominant frameworks fail to incorporate; its persistence despite available resources constitutes willful ignorance (Pohlhaus) at an institutional rather than individual level; and it reflects unacknowledged prejudices (Gadamer) that, once recognized, become contestable normative choices. The hermeneutical gap provides a diagnostic tool for identifying where ASR evaluation commits epistemic injustice -- not through technical failure but through interpretive poverty.

\subsection{From Diagnosis to Measurement}

The philosophical framework identifies \textit{what} epistemic injustice in ASR evaluation consists in. But diagnosis alone does not tell us \textit{how much} injustice occurs, \textit{where} it falls, or \textit{how} to detect it empirically. The following section develops a formal framework that operationalizes these concepts. We define a quantity that measures the evaluation cost of enforcing a single convention when others would judge the same output more favorably. The key interpretive move -- what we call the \textbf{Fricker bridge} -- is this: when this quantity differs across speaker groups, we have quantitative evidence of the structural burden that hermeneutical injustice predicts.
The relationship between EID and conventional fairness metrics is elaborated in Appendix \ref{sec:extended-related}.



\section{Formalizing the Hermeneutical Gap}


The philosophical framework developed in \S\ref{sec:philosophy} identifies hermeneutical injustice as occurring when interpretive resources are inadequate to render certain speakers' contributions intelligible. We now operationalize this concept for ASR evaluation, developing a formal framework that transforms philosophical critique into empirically tractable claims.

\subsection{Notation: Speech, Conventions, and References}

Let $\mathcal{X}$ be the space of audio segments $x$. Let $g \in G$ denote a speaker group (e.g., control/aphasia, or dialect groups such as AAE). Let $P$ be a set of \textbf{transcription policies} -- systematic choices about how to render speech as text, and $p$ is a single choice. Examples include: $p_V$ = \textit{verbatim}: preserve fillers (``um,'' ``uh''), repairs, false starts; $p_C$ = \textit{clean}: remove disfluencies, normalize to written conventions; $p_D$ = \textit{domain-specific}: transcription conventions tailored to the evidentiary or clinical requirements of a particular professional context, which may selectively combine verbatim and clean features. For example, legal transcription preserves hedges and qualifications (verbatim-like) but may normalize obvious speech errors (clean-like); medical transcription preserves symptom descriptions verbatim but normalizes filler words irrevelant to clinical documentation, and may prioritize speaker diarization results prior. Domain-specific policies are thus not simply ``more'' or ``less'' verbatim, but differently selective about which features of speech warrant preservation . 

For each policy $p \in P$, define a \textbf{reference transcript function}:
\begin{equation}
    r^{(p)}: \mathcal{X} \rightarrow \mathcal{Y}
\end{equation}
where $\mathcal{Y}$ is the space of text strings. The function $r^{(p)}$ encodes how a trained annotator following policy $p$ would transcribe audio $x$. 
Let $h: \mathcal{X} \rightarrow \mathcal{Y}$ be an ASR system (or ``hypothesis generator''), producing output $h(x)$ for input $x$.

\textbf{Key ontological move}: We do not posit a single ``true transcription'' that conventions approximate with varying fidelity. Instead, we treat the reference as \textit{constitutively} dependent on the convention. The ``ground truth'' is not discovered but \textit{constructed} through the choice of $p$.

\subsection{Plural Ground Truth}

For an utterance $x$, define the \textbf{legitimate reference set} -- where ``legitimate'' denotes any convention that is institutionally recognized, professionally practiced, and produces transcripts that serve genuine downstream purposes (as opposed to, e.g., arbitrary or adversarial renderings)\footnote{Formally, ``legitimate'' denotes any convention satisfying three conditions: (1) Institutional recognition -- the convention is codified in professional practice guidelines, annotation manuals, or domain standards. Established examples include the CHAT transcription format for clinical and child language corpora \cite{macwhinney2000childes}, the Jeffersonian transcription system conversation analysis \cite{jefferson2004glossary, sacks1974simplest}, and the court reporting standards as codified by the National Court Reporters Association \cite{ncra_cope_guidelines}. (2) Purposive coherence -- the convention serves an identifiable downstream function such that its representational choices can be justified by reference to that function. For example, verbatim preservation of disfluencies serves diagnostic purposes in clinical speech assessment, while disfluency removal serves readability for general audiences. (3) Non-adversariality -- the convention is not constructed post-hoc to minimize or maximize a particular system's score. An arbitrary rendering would be one with no coherent representational rationale (e.g., randomly deleting every third word); an adversarial rendering would be one designed specifically to make a target system appear better or worse than it is (e.g., constructing a reference that happens to match a system's known error patterns). Both are violations of (2) and (3), and are therefore excluded from $P$.}:
\begin{equation}
    R(x; P) := \left\{ r^{(p)}(x) \mid p \in P \right\}
\end{equation}

This set contains all transcriptions that could defensibly serve as ground truth given different institutionally recognized transcription policies. As previously established, the legitimacy of any element of $R(x; P)$ depends not only on professional recognition but on whose interpretive practices shaped the convention -- conventions developed without input from marginalized communities may be institutionally dominant without being epistemically authoritative for those speakers' contributions. $R(x; P)$ as defined here reflects currently practiced conventions; a fully just evaluation framework would expand $P$ to include conventions developed by or with the communities whose speech is being evaluated.

$R(x; P)$ encodes multiple ``horizons'' of transcription -- different a priori understandings about what features of speech are worth preserving in text. No element of $R(x; P)$ is privileged a priori; each represents a legitimate hermeneutical standpoint. The assumption that one element is uniquely ``correct'' reflects not objective fact but the dominance of one interpretive tradition. 

When marginalized communities develop their own transcription conventions -- as clinical speech pathologists have for aphasic speech (e.g., CHAT/CLAN coding conventions \cite{macwhinney2000childes} that mark pauses with timed intervals and code phonemic paraphasias), or as sociolinguists have for AAE (e.g., representing consonant cluster reduction or marking habitual ``be'' as a distinct morphosyntactic feature rather than as a grammatical error \cite{labov1972language, wolfram2015american, green2002african})\footnote{These conventions are grounded in descriptive sociolinguistic work establishing that AAE features are rule-governed and systematic, not errors. We note, however, that the conventions that receive institutional recognition and academic citation are often those developed by researchers studying a community rather than by community members themselves -- the descriptive versus prescriptive authority in language documentation is itself a site of epistemic injustice that our framework highlights but does not resolve.} -- these conventions constitute elements of $R(x)$ that may not be recognized by dominant evaluation frameworks. The existence of such conventions means that the interpretive resources \textit{exist}; the question is whether they are \textit{incorporated} into evaluation.

\subsection{Reference Monism Costs}

Standard ASR evaluation practice selects a single policy $p^\star$ and evaluates all systems against it:
\begin{equation}
    \text{WER}(h(x), r^{(p^\star)}(x)) = \frac{S + D + I}{N}
\end{equation}
where $S$, $D$, $I$ are the number of word substitutions, deletions, and insertions in the minimum edit sequence transforming $h(x)$ into $r^{(p^\star)}(x)$, and $N = |r^{(p^\star)}(x)|$ is the word count of the reference.

We term this practice \textbf{reference monism}: the enforcement of a single interpretive scheme as the standard against which all outputs are measured. Reference monism is not a technical necessity but a \textit{normative choice} -- one that is typically made implicitly through benchmark construction, institutional requirements, or product specifications. The cost of reference monism becomes visible when we observe that the \textit{same hypothesis} $h(x)$ receives different evaluation scores under different policies:

\begin{equation}
    \text{WER}(h(x), r^{(p_1)}(x)) \neq \text{WER}(h(x), r^{(p_2)}(x)) \quad \text{for } p_1 \neq p_2
\end{equation}

We now define a quantity that measures the cost of reference monism for a particular speaker group. Let $p^\star$ be the dominant reference policy enforced by a benchmark or institution. Define the Epistemic Injustice Distance (EID)\footnote{EID is non-negative by construction but requires care at boundary cases. If $h(x)$ is an empty string, then $\text{WER}(h(x),r^{(p)}(x))=1$ for all $p$ (all words are deletions), so $\text{EID}_g=0$ -- the monist policy imposes no additional burden, since all conventions judge the output equally harshly. This is technically correct: reference monism is not the source of harm when a system produces no output at all. The measure is designed to isolate the marginal burden of convention enforcement over and above baseline recognition failures.} for group $g$ under monist policy $p^\star$:
\begin{equation}
    \text{EID}_g(h; p^\star) := \mathbb{E}_{x \sim D_g} \left[ \text{WER}(h(x), r^{(p^\star)}(x)) - \min_{p \in P} \text{WER}(h(x), r^{(p)}(x)) \right]
\end{equation}

where $D_g$ is the distribution of audio segments from group $g$.
The expectation $\mathbb{E}_{x \sim D_g}$ is computed as the arithmetic mean over individual utterances $x$ drawn from group $g$'s data distribution $D_g$ -- that is, EID is utterance-averaged rather than computed over a single concatenated transcript. This is consistent with standard WER evaluation practice in fairness audits, where per-utterance scores are averaged across speakers within a group to avoid transcript-length artifacts \cite{mei2025addressing}.

\textbf{Interpretation}: EID measures the \textit{extra penalty} that group $g$ incurs because evaluation enforces policy $p^\star$, even when other legitimate policies in $P$ would judge the system's output more favorably. It is the cost of collapsing the legitimate reference set $R(x; P)$ to a single element $r^{(p^\star)}(x)$. 

EID has three key properties: (1) Non-negativity: $EID_g(h; p^\star) \geq 0$ always, since the enforced policy cannot perform better than the best-case policy. (2) Policy-dependence: EID depends on which convention $p^\star$ is enforced -- policies aligning with group $g$'s communicative norms yield lower EID. (3) Avoidability: If evaluation were conducted under a best-case pluralist policy -- formally, replacing the enforced $p^\star$ with $p^*_g(x)=\arg\min_{p \in P} \text{WER}(h(x), r^{(p)}(x))$ for each utterance -- then EID would be zero by construction, since the minimized term would equal the enforced term. This reveal that the burden measured by EID exists only because institutions enforce a fixed $p^\star$ ratter than permitting evaluation against the most favorable legitimate convention.



\subsection{Fricker Bridge: From Cost to Injustice}

EID measures the cost of reference monism for a single group in isolation.
To connect this individual cost to epistemic \textit{injustice}, we need a comparative structure: injustice requires that one group bears a burden that another does not, traceable to the same institutional choice. We call this the \textbf{Fricker Bridge} -- the step from measuring evaluation cost to diagnosing structural discrimination.
Define the \textbf{comparative injustice quantity} as: 
\begin{equation}
    \Delta\text{EID}(g, g'; h; p^\star) := \text{EID}_g(h; p^\star) - \text{EID}_{g'}(h; p^\star)
\end{equation}

If $\Delta\text{EID}(g, g'; h; p^\star) > 0$, then group $g$ bears a \textit{structural evaluation burden} from reference monism that group $g'$ does not bear. This burden is not merely ``performance disparity'' in the technical sense -- it is a systematic distortion of how group $g$'s speech is recorded and credited, produced by restricting the interpretive resources that evaluation recognizes.

$\Delta$EID\footnote{$\Delta$EID does not require groups $g$ and $g'$ to produce identical utterances; $D_g$ and $D_{g'}$ are different distributions over audio segments. The comparison is at the group-level expectations: we ask whether the average extra burden from reference monism differs across groups, not whether any particular pair of utterances is matched. This is analogous to how disparate impact is measured -- by comparing group-level rates, not individual-level outcomes.} operationalizes hermeneutical injustice as a measurable disparity. When $\Delta\text{EID} > 0$, group $g$'s communicative contributions are evaluated against a standard that lacks adequate interpretive resources for their speech patterns (hermeneutical injustice); group $g$ receives systematically lower credibility assessments than their speech warrants under their own community's conventions (testimonial injustice as downstream effect); and the burden falls on speakers whose conventions diverge from $p^\star$, compounding historical marginalization (Hellman's compounding injustice).

\subsection{Operationalizing the Hermeneutical Gap}

We can now give precise meaning to the hermeneutical gap concept introduced in \S \ref{sec:philosophy}. Recall that the hermeneutical gap measures the interpretive distance between a speaker's communicative contribution and the resources available in a transcription convention to render it intelligible.

The hermeneutical gap can be understood as the \textit{distance between the dominant convention and the convention most aligned with the speaker's community}. Let $p_g \in P$ be the policy that best represents group $g$'s transcription norms (e.g., verbatim clinical conventions used in speech-language pathology documentation and research, such as CHAT coding). Then:

\begin{equation}
    \mathcal{H}(g,p^\star):=\mathbb{E}_{x \sim D_g} \left[ \text{WER}(r^{(p^\star)}(x),r^{(p_g)}(x)) \right]
\end{equation}
Note that here WER is computed between two reference transcripts (measuring convention distance) rather than between a hypothesis and a reference. While WER can in principle be replaced with any string distance metric, we use WER for consistency. We note that WER is asymmetric ($\text{WER}(a,b) \neq \text{WER}(b,a)$ in general, since the denominator is the word count of the second argument); we treat $r^{(p^\star)}$ as the reference in this computation.
As with EID, the expectation is computed as the utterance-level arithmetic mean over $x \sim D_g$.

The hermeneutical gap measures convention distance \textit{independent of any ASR system}; EID measures the \textit{evaluation cost} of that gap for a particular system. A large hermeneutical gap creates the \textit{potential} for epistemic injustice; EID measures the \textit{realized} injustice when a system is evaluated under reference monism.


This relationship is not deterministic: a system specifically optimized for group $g$ under policy $p^\star$ might achieve low EID despite a large hermeneutical gap. But for systems trained on corpora reflecting majority speech demographics (e.g., LibriSpeech \cite{panayotov2015librispeech}, which draws from audiobook recordings by predominantly White, North American English speakers) and evaluated under clean, non-verbatim conventions, the hermeneutical gap $\mathcal{H}(g,p^\star)$ predicts the magnitude of EID$_g$.

\section{Empirical Evidence}

We apply the formal framework developed above to a concrete case: evaluating ASR systems on clinical speech from the AphasiaBank corpus.

\subsection{Experimental Setup}\label{sec:exp}
\subsubsection{Dataset}

We use speech samples from the AphasiaBank English Protocol dataset \cite{aphasiabank}, a corpus of semi-structured interviews with individuals with aphasia and neurotypical control participants. From the full corpus ($\sim$389 hours), we constructed a stratified test set of 8 hours 58 minutes comprising 29 control speakers and 30 speakers with aphasia (18 fluent, 12 non-fluent), sampled proportionally across clinical status, age group ($<$40, 40--59, 60--79), and gender (see Appendix \ref{app:testset} for composition details). 
Aphasic speech is characterized by disfluencies including filled pauses, false starts, word-finding delays, phonemic paraphasias, and self-repairs. These features are \textit{clinically meaningful} -- they inform diagnosis, track recovery, and characterize aphasia subtypes. Whether they should be preserved in transcription depends on the transcription's purpose.

\subsubsection{Transcription Policies}
    
We instantiate the policy set $P$ with three transcription conventions, each produced by trained human annotators (Revvers) following Rev AI's annotation guidelines: 

\begin{enumerate}
    \item $p_V$: \textbf{Verbatim} -- preserves all spoken content including fillers (``um,'' ``uh''), false starts, repetitions, and repairs. Annotators transcribe exactly what was said.
    \item $p_N$: \textbf{Non-Verbatim} -- removes disfluencies, normalizes grammar, produces ``clean'' readable text reflecting intended meaning rather than literal production.
    \item $p_L$: \textbf{Legal} -- preserves hedges, qualifications, and exact phrasing relevant for evidentiary purposes (e.g., ``I think'' vs. ``I know''), but may normalize surface speech errors such as mispronounced proper nouns or within-word false starts (e.g., ``I was go-- going'' transcribed as ``I was going'') that do not bear on meaning or intent\footnote{This reflects the clean verbatim convention used in legal transcription practices, under which epistemic and evidential markers are retained as legally significant \cite{tiersma1999legal, shuy1996language} while surface disfluencies are normalized \cite{bucholtz2000politics, fraser2022framework}. Whether a given disfluency is meaning-bearing is treated as a determination for legal professionals, not transcribers \cite{haworth2018tapes}.}.
\end{enumerate}

For each utterance $x$ in the test set, we thus have a legitimate\footnote{Legitimacy here is determined by whether the convention is professionally practiced and serves genuine purposes, not by whether the specific audio was produced in that context. A verbatim convention developed for forensic use encodes specific interpretive comments (e.g., preserving hedges) that may be equally informative when applied to clinical or everyday speech. The question of which conventions are appropriate for a given context is precisely what the dissertation makes visible, rather than presupposing.} reference set:
\begin{equation}
    R(x; P) = \left\{ r^{(p_V)}(x), r^{(p_N)}(x), r^{(p_L)}(x) \right\}
\end{equation}

All three conventions represent legitimate transcription practices used in professional contexts. None is objectively ``correct''; each serves different downstream purposes.

\subsubsection{ASR Systems}
We evaluate seven ASR configurations spanning commercial and open-source systems. We selected systems that vary along two dimensions: commercial vs. open-source architecture, and implicit vs. explicit convention control. Rev AI, unlike most ASR providers, offers multiple output modes corresponding to distinct transcription conventions: verbatim (preserving fillers, false starts, repairs), non-verbatim (clean, readable text), and legal (preserving hedges and qualifications).\footnote{Rev AI API: \url{https://www.rev.ai/}} This explicit convention control makes Rev AI uniquely suited for our analysis -- the same speech can be transcribed under different conventions by the same provider, controlling for acoustic modeling and architecture. We evaluate Rev AI v2 (verbatim, non-verbatim) and Rev AI v3 (verbatim, non-verbatim, legal), yielding five configurations.

For comparison, we include Whisper-large-v3,\footnote{\url{https://huggingface.co/openai/whisper-large-v3}} which represents widely-adopted open-source ASR and tends toward clean transcription, and CrisperWhisper,\footnote{\url{https://huggingface.co/nyrahealth/CrisperWhisper}} fine-tuned specifically to preserve disfluencies for clinical applications. Neither offers explicit convention selection; their outputs reflect implicit transcription preferences baked into training. This selection enables us to examine both explicit convention control (Rev AI modes) and implicit convention alignment (Whisper variants), demonstrating that convention-dependence affects evaluation regardless of whether systems expose it as a user-facing parameter.

\subsection{Results}

\begin{table}[h]
\centering
\caption{WER (\%) by ASR system and reference convention (overall test set). Bold indicates lowest WER for each system. Systems perform best when evaluated against ``matching'' conventions -- verbatim systems against verbatim references, etc.}
\label{tab:wer_by_convention}
\small
\begin{tabular}{lccc}
\toprule
\textbf{ASR System} & \textbf{Verbatim} & \textbf{Non-verbatim} & \textbf{Legal} \\
 & $p_V$ & $p_N$ & $p_L$ \\
\midrule
Rev AI v2 (verbatim) & \textbf{9.81} & 17.38 & 10.46 \\
Rev AI v2 (non-verbatim) & 16.18 & \textbf{9.04} & 11.46 \\
Rev AI v3 (verbatim) & \textbf{10.60} & 17.83 & 11.94 \\
Rev AI v3 (non-verbatim) & 16.95 & \textbf{9.60} & 12.71 \\
Rev AI v3 (legal) & 16.00 & 14.76 & \textbf{10.96} \\
\midrule
Whisper-large-v3 & 23.85 & \textbf{19.19} & 20.48 \\
CrisperWhisper & \textbf{29.67} & 30.65 & 26.20 \\
\bottomrule
\end{tabular}
\end{table}

Table~\ref{tab:wer_by_convention} presents WER for selected ASR systems evaluated against each reference convention.

\textbf{Key observation 1: Convention-dependence is substantial.} For Rev.ai v2 (verbatim), WER ranges from 9.81\% to 17.38\% depending solely on which reference is used -- a 1.8$\times$ difference. The same system output that appears highly accurate (9.81\%) under one convention appears substantially worse (17.38\%) under another.

\textbf{Key observation 2: Systems optimize for specific conventions}. Each system achieves its best performance against the ``matching'' reference convention: verbatim systems score lowest on verbatim references, non-verbatim systems on non-verbatim references. This is not surprising -- but it reveals that ``accuracy'' is not an intrinsic property of the system but a \textit{relational property of the system-convention pairing}.

\textbf{Key observation 3: The edit operation profile shifts.} Table~\ref{tab:edit_operations} shows how the composition of errors changes across conventions. Under verbatim reference, errors are balanced across operation types. Under non-verbatim reference, \textit{insertions} dominate (12.26\%) -- the system is penalized for producing disfluencies that the reference excludes. The same system behavior (preserving disfluencies) counts as ``correct'' under one convention and ``erroneous'' under another. \textit{What counts as an error depends on the convention.}

\begin{table}[h]
\centering
\caption{Edit operation breakdown (\%) for Rev AI v2 (verbatim) across reference conventions}
\label{tab:edit_operations}
\small
\begin{tabular}{lcccc}
\toprule
\textbf{Reference} & \textbf{WER} & \textbf{Insertions} & \textbf{Deletions} & \textbf{Substitutions} \\
\midrule
Verbatim ($p_V$) & 9.81\% & 2.18\% & 2.81\% & 4.82\% \\
Non-verbatim ($p_N$) & 17.38\% & 12.26\% & 1.30\% & 3.82\% \\
Legal ($p_L$) & 10.46\% & 4.84\% & 2.33\% & 3.30\% \\
\bottomrule
\end{tabular}
\end{table}

\subsubsection{Per Speaker Group}

\begin{table}[h]
\centering
\caption{WER (\%) by speaker group and reference convention}
\label{tab:wer_by_group}
\small
\begin{tabular}{llccc}
\toprule
\textbf{ASR System} & \textbf{Group} & \textbf{Verbatim} & \textbf{Non-verbatim} & \textbf{Legal} \\
\midrule
\multirow{3}{*}{Rev AI v2 (verbatim)} 
 & Control & 4.03 & 8.76 & 5.71 \\
 & Fluent aphasia & 14.95 & 26.06 & 13.92 \\
 & Non-fluent aphasia & 17.53 & 30.44 & 19.60 \\
\midrule
\multirow{3}{*}{Rev AI v3 (verbatim)} 
 & Control & 7.76 & 11.65 & 9.30 \\
 & Fluent aphasia & 18.85 & 28.72 & 18.72 \\
 & Non-fluent aphasia & 24.51 & 32.40 & 26.43 \\
\midrule
\multirow{3}{*}{Rev AI v3 (non-verbatim)} 
 & Control & 11.75 & 7.77 & 9.32 \\
 & Fluent aphasia & 26.26 & 17.05 & 21.68 \\
 & Non-fluent aphasia & 35.35 & 21.61 & 26.06 \\
\bottomrule
\end{tabular}
\end{table}

Table~\ref{tab:wer_by_group} disaggregates results by clinical status, revealing that convention-dependence is not uniform across groups.
Consider Rev AI v2 (verbatim):
\begin{itemize}
    \item Under verbatim reference: WER disparity between control and non-fluent aphasia groups = $17.53 - 4.03 = 13.50$ pp
    \item Under non-verbatim reference: WER disparity between control and non-fluent aphasia groups = $30.44 - 8.76 = 21.68$ pp
    \item Under legal reference: WER disparity between control and non-fluent aphasia groups = $19.60 - 5.71 = 13.89$ pp
\end{itemize}

\textbf{Key observation 4: The fairness gap depends on convention.} The ``fairness gap'' between speaker groups is not fixed -- it varies by 60\% (from 13.50 to 21.68 pp) depending on which convention is enforced. A system that appears ``moderately unfair'' under one convention appears ``severely unfair'' under another. \textit{The choice of convention determines not just absolute performance but relative fairness across groups.}

Complete WER matrices, edit operation breakdowns, per-group results for all systems, and inter-reference distance calculations operationalizing the hermeneutical gap are provided in Appendix~\ref{app:experiment}.

\subsection{Computing EID}
We now compute EID for each speaker group, demonstrating how reference monism imposes differential burdens.
We treat non-verbatim transcription ($p_N$) as the dominant enforced policy $p^\star$, reflecting common evaluation practice: major ASR benchmarks including LibriSpeech \cite{panayotov2015librispeech}, CommonVoice \cite{ardila2020common}, and FLEURS \cite{conneau2023fleurs} all use normalized, clean references with disfluencies removed or absent by design. This convention is so prevalent it is rarely stated explicitly -- benchmarks report ``WER'' and not ``WER under clean conventions,'' naturalizing the choice and rendering it invisible.
For each group $g$, we compute:
\begin{equation}
    \text{EID}_g(h; p_N) = \text{WER}(h, r^{(p_N)}) - \min_{p \in \{p_V, p_N, p_L\}} \text{WER}(h, r^{(p)})
\end{equation}

Consider Rev AI v2 (verbatim) as an illustrative case. Under non-verbatim evaluation, control speakers achieve 8.76\% WER against the enforced reference but only 4.03\% against their best-case reference (verbatim), yielding EID of 4.73 pp. Fluent aphasic speakers show 26.06\% WER under enforcement versus 13.92\% at best (legal), for EID of 12.14 pp. Non-fluent aphasic speakers fare worst: 30.44\% WER under enforcement versus 17.53\% at best (verbatim), yielding EID of 12.91 pp -- nearly three times the burden borne by control speakers.

Computing $\Delta$EID:

\begin{align}
    \Delta\text{EID}(\text{non-fluent}, \text{control}; h; p_N) = 12.91 - 4.73 = 8.18 \text{ pp}
\end{align}

\begin{figure}
    \centering
    \includegraphics[width=\textwidth]{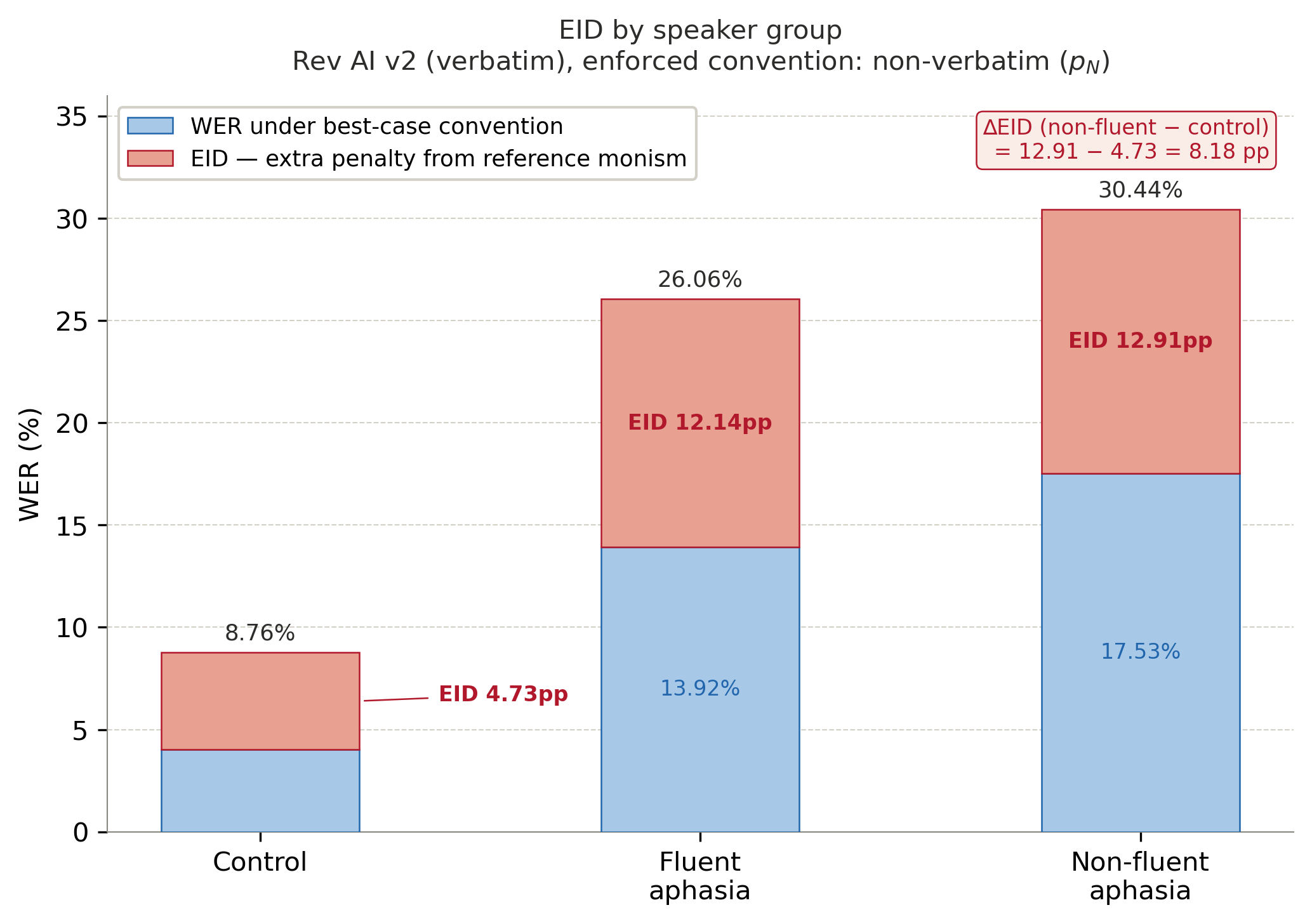}
    \caption{EID decomposition by speaker group for Rev AI v2 (verbatim) under enforced non-verbatim evaluation ($p^\star = p_N$). Each bar shows WER under the best-case convention (teal, bottom) plus the EID penalty from reference monism (coral, top), which together sum to WER under the enforced convention. The coral portion grows disproportionately for aphasic speakers: non-fluent aphasic speakers bear an EID of 12.91pp versus 4.73pp for control speakers, yielding $\Delta\text{EID} = 8.18$pp. This penalty is not a property of the speakers or the ASR system, but of the institutional choice to enforce a single transcription convention on speech characterized by clinically meaningful disfluency.}
    \label{fig:eid-groups}
\end{figure}

\textbf{Normative reading}: Speakers with non-fluent aphasia bear an additional 8.18 percentage point penalty from reference monism compared to control speakers. This is the \textit{structural evaluation burden} -- not a performance disparity in the usual sense, but a systematic penalty imposed by the choice to enforce non-verbatim conventions on speech that is inherently disfluent. Figure~\ref{fig:eid-groups} visualizes this decomposition: each bar shows WER under the best-case convention (teal) plus the EID penalty imposed by reference monism (coral), with the coral portion growing disproportionately for aphasic speakers.

\begin{table}[h]
\centering
\caption{EID by speaker group and $\Delta$EID across ASR systems (enforced policy: non-verbatim). EID in percentage points (pp). Higher EID indicates greater structural burden from reference monism. Positive $\Delta$EID indicates non-fluent speakers bear greater burden than control speakers.}
\label{tab:eid}
\small
\begin{tabular}{lcccc}
\toprule
\textbf{ASR System} & \textbf{EID (Control)} & \textbf{EID (Fluent)} & \textbf{EID (Non-fluent)} & \textbf{$\Delta$EID} \\
\midrule
Rev AI v2 (verbatim) & 4.73 pp & 12.14 pp & 12.91 pp & 8.18 pp \\
Rev AI v3 (verbatim) & 3.89 pp & 9.87 pp & 7.89 pp & 4.00 pp \\
Rev AI v3 (non-verbatim) & 0.00 pp & 0.00 pp & 0.00 pp & 0.00 pp \\
Rev AI v3 (legal) & 4.89 pp & 6.69 pp & 3.52 pp & $-$1.37 pp \\
\bottomrule
\end{tabular}
\end{table}

Table~\ref{tab:eid} shows EID and $\Delta$EID across multiple systems. \textbf{Key observation 5: EID depends on system-convention alignment.} When the ASR system is optimized for the enforced convention (Rev AI v3 non-verbatim evaluated against non-verbatim reference), EID is zero by construction -- the enforced convention \textit{is} the best convention for that system. But for systems optimized for other conventions, substantial EID emerges, and it falls disproportionately on aphasic speakers\footnote{The negative $\Delta$EID for Rev AI v3 (legal) ($-$1.37pp) warrants explanation. Under this system, the legal reference $p_L$ is the \textit{best-case} convention for control speakers, because Rev AI v3 (legal) is calibrated to preserve hedges and qualifications -- features more characteristic of fluent, carefully hedged control speech. The enforced non-verbatim convention $p_N$ therefore penalizes control speakers more than it penalizes non-fluent aphasic speakers, who produce shorter, less hedged utterances for which verbatim remains the best-case convention. Consequently, EID for control speakers exceeds EID for non-fluent aphasic speakers under this system, reversing the typical direction of $\Delta$EID. This illustrates that $\Delta$EID is system-convention specific: it does not always favor the historically privileged group, but reflects the particular alignment between a system's output tendencies and the enforced convention.}.

\subsection{Fricker Bridge, Empirically}

The positive $\Delta$EID confirms our theoretical prediction: enforcing a single transcription convention commits hermeneutical injustice against speakers whose communicative practices diverge from that convention.

\begin{itemize}
    \item \textbf{Hermeneutical ghettoization}: Clinical speech communities have developed transcription conventions that preserve disfluencies as meaningful (verbatim transcription is standard in clinical research). These conventions exist in $P$ but are excluded when $p_N$ is enforced as the evaluation standard.
    
    \item \textbf{Willful hermeneutical ignorance}: The interpretive resources to fairly evaluate aphasic speech \textit{exist} -- verbatim conventions produce dramatically lower WER for these speakers. The injustice arises from institutional practices that enforce non-verbatim conventions as the unmarked default.
    
    \item \textbf{The evaluation cost is quantifiable}: $\Delta$EID of 8.18 pp means that non-fluent aphasic speakers are penalized by an additional 8 percentage points purely due to convention choice -- a penalty that would disappear under pluralist evaluation.
\end{itemize}

\section{Reporting with WER-Range}\label{sec:wer-range}
The results above demonstrate that no single WER number adequately characterizes system performance. The ``accuracy'' of a system is not a fact but an artifact of convention choice. We propose an alternative: rather than collapsing the legitimate reference set to a single ground truth, \textbf{report the range of WER values across legitimate conventions}.

For an ASR system $h$ evaluated on dataset $D$ with policy set $P$, define the \textbf{WER-Range}:
\begin{equation}
    \text{WER-Range}(h, D, P) := \left[ \min_{p \in P} \text{WER}_p, \; \max_{p \in P} \text{WER}_p \right]
\end{equation}
Equivalently, report the \textbf{WER-Range}:
\begin{equation}
    \text{WER-Set}(h, D, P) := \left\{ (p, \text{WER}_p) : p \in P \right\}
\end{equation}

Instead of reporting:
\begin{quote}
    ``Rev AI v2 (verbatim) achieves 9.81\% WER on AphasiaBank.''
\end{quote}
Report:
\begin{quote}
    ``Rev AI v2 (verbatim) achieves \textbf{WER-Range [9.81\%, 17.38\%]} on AphasiaBank across verbatim, non-verbatim, and legal transcription conventions.''
\end{quote}
Or more informatively:
\begin{quote}
    ``Rev AI v2 (verbatim) achieves WER-Set \{9.81\% (verbatim), 10.46\% (legal), 17.38\% (non-verbatim)\} on AphasiaBank.''
\end{quote}

\begin{table}[h]
\centering
\caption{WER-Range by ASR system (top) and by speaker group for Rev AI v2 verbatim (bottom). Range width measures vulnerability to convention choice; wider ranges indicate greater sensitivity to which standard defines ground truth.}
\label{tab:wer_range}
\small
\begin{tabular}{lcc}
\toprule
\textbf{ASR System} & \textbf{WER-Range} & \textbf{Range Width} \\
\midrule
Rev AI v2 (verbatim) & [9.81\%, 17.38\%] & 7.57 pp \\
Rev AI v2 (non-verbatim) & [9.04\%, 16.18\%] & 7.14 pp \\
Rev AI v3 (verbatim) & [10.60\%, 17.83\%] & 7.23 pp \\
Rev AI v3 (non-verbatim) & [9.60\%, 16.95\%] & 7.35 pp \\
Rev AI v3 (legal) & [10.96\%, 16.00\%] & 5.04 pp \\
Whisper-large-v3 & [19.19\%, 23.85\%] & 4.66 pp \\
CrisperWhisper & [26.20\%, 30.65\%] & 4.45 pp \\
\midrule
\textbf{Speaker Group} & \textbf{WER-Range} & \textbf{Range Width} \\
\midrule
Control & [4.03\%, 8.76\%] & 4.73 pp \\
Fluent aphasia & [13.92\%, 26.06\%] & 12.14 pp \\
Non-fluent aphasia & [17.53\%, 30.44\%] & 12.91 pp \\
\bottomrule
\end{tabular}
\end{table}

Table~\ref{tab:wer_range} presents WER-Range for all evaluated systems and, for Rev AI v2 (verbatim), disaggregated by speaker group. Range width varies across systems: Rev AI v3 (legal) shows the narrowest range among Rev systems (5.04 pp), suggesting more robust performance across conventions, while CrisperWhisper shows the narrowest absolute range (4.45 pp) but at higher overall WER -- it performs similarly (poorly) across all conventions.

Disaggregating by speaker group reveals differential vulnerability to convention choice. Non-fluent aphasic speakers have a WER-Range width of 12.91 pp -- nearly three times the width of control speakers (4.73 pp). This means aphasic speakers are \textit{more vulnerable to convention choice}: their apparent ``accuracy'' fluctuates more depending on which standard is enforced. This differential vulnerability is invisible to single-number reporting but captured by WER-Range. Note that range width equals EID when non-verbatim is the worst-case convention -- which it is for verbatim-optimized systems evaluating disfluent speech.

We address potential objections -- including why WER averaging or cross-dataset evaluation don't serve the same purpose -- in Appendix \ref{sec:alternative-approaches}


\section{Decomposing Evaluation Costs}

Does plural ground truth impose prohibitive costs? The concern is legitimate, but the cost objection conflates two distinct issues -- the practical question of resource allocation and the epistemic question of what counts as correct transcription.

Because our proposal concerns evaluation infrastructure rather than runtime transcription, costs are incurred once during benchmark construction rather than per-deployment.
ASR evaluation incurs two cost categories that respond differently to plural ground truth. \textbf{Computational costs} (inference, WER computation, analysis) are \textit{invariant to reference multiplicity}: whether we evaluate against one reference or three, we run inference once per audio segment, so inference cost is $O(MN)$. WER computation scales with reference multiplicity, giving total evaluation cost $O(MN|P|)$, where the $|P|$ factor is computationally trivial. \textbf{Human annotation costs} (transcriber time, quality assurance) scale with audio duration times policies: $O(D \cdot |P|)$. This is where plural ground truth increases expenses.

\subsection{Mitigating Annotation Costs}

Three factors reduce the practical burden. First, \textbf{benchmark amortization}: plural references are evaluation infrastructure, produced once and reused across all subsequent system assessments. A benchmark with plural ground truth references enable unlimited evaluations at zero marginal annotation cost. Moreover, systematic yet simple post-processing rules can generate convention variants from a single high-fidelity verbatim transcript reducing annotation to one careful pass per utterance. Second, \textbf{labor market expansion}: multi-reference annotation sustains professional transcription expertise -- clinical transcriptionists, legal transcribers, sociolinguistic annotators -- rather than replacing it with under-specified ``general purpose'' annotation. Third, \textbf{selective deployment}: narrow WER-Range indicates convention choice matters little, justifying single-reference reporting; wide WER-Range signals plural evaluation is necessary. Initial multi-reference evaluation can diagnose when cheaper approaches suffice.

\subsection{Who Should Bear These Costs?}

Plural ground truth is an evaluation methodology concerning how we assess systems, not how end users interact with them. \textbf{Academic researchers} face genuine resource constraints, but research evaluation serves a gatekeeping function -- misspecified benchmarks propagate errors downstream. \textbf{Commercial deployers} have direct financial interest in accurate evaluation; the cost of multi-reference evaluation is negligible compared to deploying an ill-suited system. \textbf{Technology companies} possess resources for comprehensive evaluation but often lack incentive. We make a normative claim: companies marketing speech technologies as universally accessible have an obligation to evaluate against interpretive frameworks used by the communities they claim to serve.

The cost objection obscures a deeper issue. There is a categorical difference between \textit{epistemic humility} -- ``We recognize multiple legitimate interpretations exist, but resource constraints prevent evaluating all; we report WER under convention $p^*$ while acknowledging this limitation'' -- and \textit{epistemic monism} -- ``We evaluate against $p^*$ because it represents the true transcription.'' Current practice embodies the latter. Benchmarks report ``WER,'' not ``WER under clean conventions.'' The convention is naturalized, rendered invisible. Even when constraints permit only one convention, reporting ``WER = 15\% (non-verbatim)'' rather than ``WER = 15\%'' acknowledges that accuracy is relational. This costs nothing and changes everything.

For organizations committed to fair evaluation despite resource constraints, we provide a staged implementation roadmap in Appendix \ref{sec:roadmap}.

\section{Conclusion}
This paper has argued that standard ASR evaluation commits epistemic injustice -- not through technical failure but through interpretive poverty. By enforcing a single transcription convention as ground truth, reference monism systematically disadvantages speakers whose communicative practices diverge from that convention. Our empirical results show that WER for identical ASR output varies by nearly a factor of two depending on which convention defines ground truth, and that this variation falls disproportionately on speakers with aphasia.

The practical upshot is WER-Range: report the range of accuracy values across legitimate transcription conventions rather than a single number. This shift does not resolve all fairness concerns, but it distinguishes two sources of disparity that reference monism conflates -- those arising from system limitations and those arising from evaluative infrastructure. Only the former are technical problems; the latter are normative choices masquerading as measurement.

Our analysis has limitations. Empirically, we demonstrate plural ground truth using a single corpus (AphasiaBank), three transcription conventions, and one dimension of speaker variation (clinical status). While the theoretical framework applies wherever multiple legitimate conventions exist -- including dialectal variation, non-native speech, child language, and accented speech -- we have not validated EID and $\Delta$EID for these populations, and our convention set could be expanded to include medical, linguistic, or accessibility-focused transcription standards. Future work should also assess statistical robustness through bootstrapping or mixed-effects modeling given the relatively small speaker counts per group.

Clinical speech communities and sociolinguists have developed transcription conventions adequate to marginalized speakers' communicative practices. These interpretive resources exist. Continued reliance on reference monism despite their availability constitutes willful hermeneutical ignorance. The cost of producing multiple references is real; the cost of assuming only one correct reference -- when that assumption predictably disadvantages marginalized communities -- is epistemic injustice. We have tried to make both visible.

\newpage

\appendix

\section{Extended Philosophical Background}\label{app:philosophy}

This appendix elaborates the philosophical concepts introduced in \S~\ref{sec:philosophy}.

\subsection{Extensions of Testimonial Injustice}

Davis \cite{davis2016typecasts} extends Fricker's analysis to \textit{credibility excess}: being judged \textit{more} credible due to positive stereotypes also constitutes testimonial injustice, because the speaker is treated as a representative of their group rather than as an individual epistemic agent. This bidirectional analysis proves relevant for ASR: speakers of ``standard'' dialects receive credibility excess -- their speech patterns assumed correct by default -- while speakers of marginalized dialects receive credibility deficit.

\subsection{Goetze's Taxonomy of Hermeneutical Injustice}

Goetze \cite{goetze2018hermeneutical} provides a taxonomy of six species of hermeneutical injustice, distinguished by who possesses the relevant interpretive tools (Table~\ref{tab:species_appendix}).

\begin{table}[h]
\centering
\caption{Species of hermeneutical injustice, adapted from Goetze \cite{goetze2018hermeneutical}}
\label{tab:species_appendix}
\small
\begin{tabular}{lcccc}
\toprule
\textbf{Species} & \textbf{Subject} & \textbf{Subject's group} & \textbf{Other groups} & \textbf{Primary harm} \\
\midrule
Effacement & No & No & No & Cognitive \\
Isolation & Yes & No & No & Communicative \\
Separation & No & No & Yes & Cognitive \\
Ghettoization & Yes & Yes & No & Communicative \\
Exportation & Yes & No & Yes & Communicative \\
Obstruction & Yes & Yes & Yes (some) & Communicative \\
\bottomrule
\end{tabular}
\end{table}

The taxonomy reveals that hermeneutical injustice manifests as either \textit{cognitive} harm (the subject cannot understand her own experience) or \textit{communicative} harm (the subject understands but cannot make others understand). Both are instances of the same fundamental wrong: at some crucial moment, the subject's experience lacks intelligibility due to gaps in available interpretive resources.

For ASR, \textit{hermeneutical ghettoization} is most salient: clinical speech communities have developed verbatim transcription conventions preserving disfluencies as clinically meaningful; AAE-speaking communities have well-documented linguistic systems -- including phonological features such as consonant cluster reduction and syllable-initial fricative stopping, morphosyntactic patterns such as habitual ``be'' and copula deletion, and prosodic conventions -- that are fully described in their sociolinguistic literature \cite{labov1972language, rickford2007spoken}. These interpretive resources exist within these communities but have not been incorporated into mainstream ASR evaluation frameworks -- leaving speakers unable to ``communicate'' their speech to systems that lack the interpretive tools to recognize it as legitimate.

\subsection{Dreyfus's Critique: Three Dimensions}

Three aspects of Dreyfus \cite{dreyfus1992computers}'s critique apply to ASR ground truth:

\textbf{The frame problem.} Determining what counts as ``relevant'' context requires appealing to larger contexts, leading to infinite regress. What counts as ``correct'' transcription depends on purpose -- legal documentation, medical records, linguistic research -- but specifying which purpose is relevant requires further contextual judgment. The assumption that transcription can be evaluated against purpose-independent ground truth ignores this fundamental dependence.

\textbf{Tacit knowledge.} Expert transcribers operate through intuitive expertise that cannot be articulated as explicit rules. When professionals disagree about rendering an utterance, they are exercising different tacit understandings of what matters in context. Aggregating judgments into ``ground truth'' yields consensus -- a social construction reflecting the perspectives of those who predominate in transcription professions -- not objectivity.

\textbf{Embodied understanding.} Dreyfus emphasized that human understanding is fundamentally embodied and situated. When humans comprehend speech, we interpret communicative acts embedded in social situations, speaker relationships, and shared backgrounds. A transcriber hearing a speaker pause, restart, and rephrase understands this as word-finding difficulty, nervousness, or emphasis depending on context. ASR systems process what Dreyfus called ``isolated domains'' of acoustic features, stripped of embodied context. They miss what Dreyfus termed ``solicitations'' -- the ways environment calls forth appropriate responses without explicit reasoning.

fjelland \cite{fjelland2020general} recently revived Dreyfus's critique for deep learning, arguing that even modern neural systems cannot handle genuinely novel situations requiring contextual judgment. For ASR, no amount of training data substitutes for the interpretive flexibility human transcribers bring -- flexibility excluded when judgments are flattened into fixed labels.

\subsection{Gadamer's Hermeneutic Circle}

\citet{gadamer2013truth}'s concept of the \textit{hermeneutic circle} further challenges ASR's approach. The circle describes a fundamental structure: we cannot understand parts of an utterance without grasping the whole, yet we grasp the whole only through understanding parts. This is productive, not vicious -- interpretation proceeds by moving between part and whole, revising understanding of each.

Human transcribers embody this movement. Hearing an ambiguous word, they interpret it in light of the sentence; understanding the sentence, they revise their sense of the word. As conversation unfolds, they continuously adjust interpretation. A word that seemed erroneous may reveal itself as intentional repetition; a pause that seemed like disfluency may emerge as emphasis.

ASR systems typically process speech in segments, committing to local decisions without capacity to revise earlier interpretations as context emerges -- lacking the circular, revisionary structure that makes human understanding possible.

\subsection{Synthesis: Seven Philosophical Resources}

The hermeneutical gap synthesizes seven philosophical contributions:

\begin{itemize}
    \item From \textbf{Fricker}: the gap constitutes hermeneutical injustice when it results from the speaker's marginalization from processes that shaped transcription conventions.
    \item From \textbf{Goetze}: when marginalized communities have developed their own conventions (hermeneutical dissent), the gap reflects ghettoization -- dominant frameworks' failure to incorporate available interpretive resources.
    \item From \textbf{Dotson}: the gap induces both testimonial quieting (speech is misrecognized) and testimonial smothering (speakers modify their voice to close the gap).
    \item From \textbf{Anderson}: the gap is structural, embedded in institutions of transcription and evaluation, not remediable through individual system improvements alone.
    \item From \textbf{Pohlhaus}: the persistence of the gap despite available resources constitutes willful hermeneutical ignorance.
    \item From \textbf{Dreyfus}: the assumption that the gap can be closed by identifying ``true'' ground truth misunderstands the contextual, tacit nature of transcription expertise.
    \item From \textbf{Gadamer}: the gap reflects unacknowledged prejudices in transcription conventions that, once recognized, become contestable normative choices.
\end{itemize}

\section{Extended Experimental Results}\label{app:experiment}

This appendix provides complete experimental data summarized in \S~\ref{sec:exp}.

\subsection{Test Set Composition}\label{app:testset}

Our test set comprises 8 hours 58 minutes of speech from the AphasiaBank English Protocol dataset, stratified by clinical status, age group, and gender. 
Table~\ref{tab:balance} summarizes balance across primary grouping variables.


\begin{table}[h]
\centering
\caption{Test set balance verification}
\label{tab:balance}
\small
\begin{tabular}{llcc}
\toprule
\textbf{Variable} & \textbf{Level} & \textbf{Files} & \textbf{Hours} \\
\midrule
\multirow{3}{*}{Clinical Status} & Control & 29 & 3.13 \\
 & Fluent Aphasia & 18 & 3.55 \\
 & Non-fluent Aphasia & 12 & 2.30 \\
\midrule
\multirow{2}{*}{Gender} & Female & 30 & 4.10 \\
 & Male & 29 & 4.87 \\
\midrule
\multirow{3}{*}{Age Group} & 60--79 & 33 & 5.34 \\
 & 40--59 & 21 & 3.21 \\
 & $<$40 & 5 & 0.42 \\
\bottomrule
\end{tabular}
\end{table}




\subsection{Edit Operation Breakdown}\label{app:edit_ops}

Table~\ref{tab:full_edit_ops} presents the complete breakdown of WER into insertions (I), deletions (D), and substitutions (S) for all Rev AI systems across all reference conventions.

\begin{table}[h]
\centering
\caption{Edit operation breakdown (\%) by ASR system and reference convention}
\label{tab:full_edit_ops}
\small
\begin{tabular}{llcccc}
\toprule
\textbf{ASR System} & \textbf{Reference} & \textbf{WER} & \textbf{I} & \textbf{D} & \textbf{S} \\
\midrule
\multirow{3}{*}{Rev AI v2 (verbatim)} 
 & Verbatim & 9.81 & 2.18 & 2.81 & 4.82 \\
 & Non-verbatim & 17.38 & 12.26 & 1.30 & 3.82 \\
 & Legal & 10.46 & 4.84 & 2.33 & 3.30 \\
\midrule
\multirow{3}{*}{Rev AI v2 (non-verbatim)} 
 & Verbatim & 16.18 & 1.03 & 11.34 & 3.80 \\
 & Non-verbatim & 9.04 & 3.17 & 2.43 & 3.44 \\
 & Legal & 11.46 & 0.68 & 8.19 & 2.59 \\
\midrule
\multirow{3}{*}{Rev AI v3 (verbatim)} 
 & Verbatim & 10.60 & 2.27 & 3.35 & 4.98 \\
 & Non-verbatim & 17.83 & 12.11 & 1.66 & 4.05 \\
 & Legal & 11.94 & 4.97 & 2.93 & 4.04 \\
\midrule
\multirow{3}{*}{Rev AI v3 (non-verbatim)} 
 & Verbatim & 16.95 & 0.98 & 12.06 & 3.91 \\
 & Non-verbatim & 9.60 & 2.65 & 3.41 & 3.53 \\
 & Legal & 12.71 & 0.75 & 9.06 & 2.89 \\
\midrule
\multirow{3}{*}{Rev AI v3 (legal)} 
 & Verbatim & 16.00 & 2.45 & 6.21 & 7.34 \\
 & Non-verbatim & 14.76 & 8.74 & 1.32 & 4.71 \\
 & Legal & 10.96 & 1.68 & 2.43 & 6.85 \\
\bottomrule
\end{tabular}
\end{table}

\textbf{Interpretation}: The edit operation profile systematically shifts with convention mismatch:
\begin{itemize}
    \item \textbf{Verbatim ASR $\times$ Non-verbatim reference}: Insertions dominate (12.26\%, 12.11\%). The system produces disfluencies that the reference excludes, so preserved fillers become ``spurious insertions.''
    \item \textbf{Non-verbatim ASR $\times$ Verbatim reference}: Deletions dominate (11.34\%, 12.06\%). The system removes disfluencies that the reference preserves, so omitted fillers become ``missing words.''
    \item \textbf{Matched systems}: Edit operations are balanced (I $\approx$ D $\approx$ S), reflecting genuine transcription errors rather than convention mismatch.
\end{itemize}

This pattern demonstrates that \textit{what counts as an error} is convention-dependent. The same system behavior -- preserving or removing disfluencies -- registers as correct or erroneous depending solely on the reference convention.

\subsection{Per-Group Results for All Systems}\label{app:per_group}

Table~\ref{tab:group_verbatim} presents WER disaggregated by clinical status for all ASR systems, under each reference convention, and table~\ref{tab:fairness_gaps} summarizes the fairness gap (non-fluent $-$ control) across all system-reference combinations.

\begin{table}[h]
\centering
\caption{WER (\%) by speaker group, Reference: Verbatim}
\label{tab:group_verbatim}
\small
\begin{tabular}{lcccc}
\toprule
\textbf{ASR System} & \textbf{Control} & \textbf{Aphasia (all)} & \textbf{Fluent} & \textbf{Non-fluent} \\
\midrule
Rev AI v2 (verbatim) & 4.03 & 15.67 & 14.95 & 17.53 \\
Rev AI v2 (non-verbatim) & 8.26 & 24.22 & 22.74 & 28.10 \\
Rev AI v3 (verbatim) & 7.76 & 20.44 & 18.85 & 24.51 \\
Rev AI v3 (non-verbatim) & 11.75 & 28.82 & 26.26 & 35.35 \\
Rev AI v3 (legal) & 10.83 & 27.61 & 25.59 & 32.76 \\
\bottomrule
\end{tabular}
\end{table}



\begin{table}[h]
\centering
\caption{Fairness gap (WER\textsubscript{non-fluent} $-$ WER\textsubscript{control}) in percentage points}
\label{tab:fairness_gaps}
\small
\begin{tabular}{lccc}
\toprule
\textbf{ASR System} & \textbf{Verbatim} & \textbf{Non-verbatim} & \textbf{Legal} \\
\midrule
Rev AI v2 (verbatim) & 13.50 & 21.68 & 13.89 \\
Rev AI v2 (non-verbatim) & 19.84 & 10.72 & 11.27 \\
Rev AI v3 (verbatim) & 16.75 & 20.75 & 17.13 \\
Rev AI v3 (non-verbatim) & 23.60 & 13.84 & 16.74 \\
Rev AI v3 (legal) & 21.93 & 15.26 & 14.45 \\
\bottomrule
\end{tabular}
\end{table}

\textbf{Key observation}: The fairness gap varies by 50--100\% depending on reference convention. For Rev AI v2 (verbatim), the gap ranges from 13.50 pp (verbatim reference) to 21.68 pp (non-verbatim reference) -- a 60\% increase. For Rev AI v2 (non-verbatim), the pattern reverses: the gap is largest under verbatim reference (19.84 pp) and smallest under non-verbatim reference (10.72 pp). \textit{Which system appears ``fairer'' depends entirely on which convention defines ground truth.}

\subsection{Inter-Reference Distance}\label{app:interref}

To directly measure how much transcription conventions diverge, we compute the edit distance between reference pairs using each ASR hypothesis as an anchor. This operationalizes the hermeneutical gap: the distance a speaker's contribution must ``travel'' when evaluated under a convention misaligned with their communicative norms.

For each ASR output $h(x)$, we identify which words match each reference, then compute WER between reference pairs based on alignment. Table~\ref{tab:interref} presents results.

\begin{table}[h]
\centering
\caption{Inter-reference distance: WER (\%) between reference pairs, by ASR anchor}
\label{tab:interref}
\small
\begin{tabular}{llcccc}
\toprule
\textbf{ASR Anchor} & \textbf{Reference Pair} & \textbf{WER (std)} & \textbf{I} & \textbf{D} & \textbf{S} \\
\midrule
\multirow{3}{*}{Rev AI v2 (verbatim)} 
 & Legal -- Verbatim & 5.71 (5.65) & 1.81 & 1.58 & 2.32 \\
 & Legal -- Non-verbatim & 9.01 (7.81) & 5.46 & 1.27 & 2.29 \\
 & Verbatim -- Non-verbatim & 7.43 (8.12) & 3.06 & 1.36 & 3.01 \\
\midrule
\multirow{3}{*}{Rev AI v2 (non-verbatim)} 
 & Legal -- Verbatim & 9.55 (6.61) & 0.72 & 6.86 & 1.98 \\
 & Legal -- Non-verbatim & 5.72 (5.10) & 1.15 & 2.63 & 1.95 \\
 & Verbatim -- Non-verbatim & 6.81 (6.65) & 1.61 & 2.58 & 2.63 \\
\midrule
\multirow{3}{*}{Rev AI v3 (verbatim)} 
 & Legal -- Verbatim & 6.79 (6.65) & 2.09 & 2.03 & 2.67 \\
 & Legal -- Non-verbatim & 10.10 (8.00) & 5.91 & 1.56 & 2.64 \\
 & Verbatim -- Non-verbatim & 7.92 (7.87) & 3.29 & 1.53 & 3.09 \\
\midrule
\multirow{3}{*}{Rev AI v3 (non-verbatim)} 
 & Legal -- Verbatim & 10.52 (10.09) & 0.78 & 7.61 & 2.13 \\
 & Legal -- Non-verbatim & 6.57 (8.68) & 1.19 & 3.20 & 2.18 \\
 & Verbatim -- Non-verbatim & 7.39 (9.21) & 1.53 & 3.25 & 2.61 \\
\midrule
\multirow{3}{*}{Rev AI v3 (legal)} 
 & Legal -- Verbatim & 9.39 (7.25) & 1.96 & 1.79 & 5.63 \\
 & Legal -- Non-verbatim & 9.29 (7.12) & 3.92 & 1.11 & 4.27 \\
 & Verbatim -- Non-verbatim & 10.91 (8.90) & 5.25 & 1.30 & 4.36 \\
\bottomrule
\end{tabular}
\end{table}

\textbf{Interpretation}: The distance between verbatim and non-verbatim references ranges from 6.81\% to 10.91\% depending on the ASR anchor used for alignment. This represents the \textit{irreducible divergence} between conventions -- the portion of ``error'' attributable to interpretive framework choice rather than system performance.

Three patterns emerge:
\begin{enumerate}
    \item \textbf{Verbatim--Non-verbatim distance is substantial}: Averaging across anchors, verbatim and non-verbatim references differ by approximately 8\% WER. This is the hermeneutical gap in quantitative terms.
    
    \item \textbf{Legal occupies middle ground}: Legal--Verbatim distance (5.71--10.52\%) and Legal--Non-verbatim distance (5.72--10.10\%) vary depending on the ASR anchor, suggesting legal transcription shares features with both extremes.
    
    \item \textbf{Edit operations reveal convention semantics}: The Verbatim--Non-verbatim distance is dominated by insertions when the anchor is verbatim-oriented (non-verbatim reference lacks the disfluencies) and by deletions when the anchor is non-verbatim-oriented (verbatim reference has ``extra'' words). This confirms that the distance reflects systematic convention differences, not random variation.
\end{enumerate}

This inter-reference distance bounds the possible improvement from system optimization alone: even a ``perfect'' system cannot score below the convention distance when evaluated against a misaligned reference. For speakers whose natural communicative patterns align with verbatim conventions but who are evaluated against non-verbatim references, this 7--11\% distance represents an irreducible penalty -- the quantified hermeneutical gap.

\subsection{Extended EID Calculations}\label{app:name}

Table~\ref{tab:full_name} presents EID calculations for all systems and speaker groups, under both non-verbatim and verbatim enforcement scenarios.

\begin{table}[h]
\centering
\caption{EID by speaker group across all systems}
\label{tab:full_name}
\small
\begin{tabular}{llcccc}
\toprule
& & \multicolumn{2}{c}{\textbf{Enforced: Non-verbatim}} & \multicolumn{2}{c}{\textbf{Enforced: Verbatim}} \\
\cmidrule(lr){3-4} \cmidrule(lr){5-6}
\textbf{ASR System} & \textbf{Group} & \textbf{EID} & \textbf{Best $p$} & \textbf{EID} & \textbf{Best $p$} \\
\midrule
\multirow{3}{*}{Rev AI v2 (verbatim)} 
 & Control & 4.73 & V & 0.00 & V \\
 & Fluent & 12.14 & L & 1.03 & L \\
 & Non-fluent & 12.91 & V & 0.00 & V \\
\midrule
\multirow{3}{*}{Rev AI v2 (non-verb.)} 
 & Control & 0.00 & N & 3.68 & N \\
 & Fluent & 0.00 & N & 9.04 & N \\
 & Non-fluent & 0.00 & N & 12.80 & N \\
\midrule
\multirow{3}{*}{Rev AI v3 (verbatim)} 
 & Control & 3.89 & V & 0.00 & V \\
 & Fluent & 9.87 & L & 0.13 & L \\
 & Non-fluent & 7.89 & V & 0.00 & V \\
\midrule
\multirow{3}{*}{Rev AI v3 (non-verb.)} 
 & Control & 0.00 & N & 3.98 & N \\
 & Fluent & 0.00 & N & 9.21 & N \\
 & Non-fluent & 0.00 & N & 13.74 & N \\
\midrule
\multirow{3}{*}{Rev AI v3 (legal)} 
 & Control & 4.89 & L & 2.48 & L \\
 & Fluent & 6.69 & L & 6.32 & L \\
 & Non-fluent & 3.52 & L & 9.96 & L \\
\bottomrule
\end{tabular}%
\end{table}

\textbf{Key patterns}:
\begin{itemize}
    \item Systems achieve EID = 0 when the enforced policy matches their optimization target (e.g., non-verbatim systems under non-verbatim enforcement).
    \item For verbatim-optimized systems under non-verbatim enforcement, aphasic speakers bear 2--3$\times$ higher EID than control speakers.
    \item For non-verbatim-optimized systems under verbatim enforcement, the pattern reverses but remains: aphasic speakers bear higher EID.
    \item The ``best policy'' column reveals that fluent aphasic speakers sometimes benefit most from legal conventions (which preserve some but not all disfluencies), while non-fluent aphasic speakers consistently benefit from verbatim conventions.
\end{itemize}

Table~\ref{tab:full_delta_name} presents $\Delta$EID across all systems.

\begin{table}[h]
\centering
\caption{$\Delta$EID (Non-fluent $-$ Control) across systems and enforcement scenarios}
\label{tab:full_delta_name}
\small
\begin{tabular}{lcc}
\toprule
\textbf{ASR System} & \textbf{Enforced: Non-verbatim} & \textbf{Enforced: Verbatim} \\
\midrule
Rev AI v2 (verbatim) & +8.18 pp & 0.00 pp \\
Rev AI v2 (non-verbatim) & 0.00 pp & +9.12 pp \\
Rev AI v3 (verbatim) & +4.00 pp & 0.00 pp \\
Rev AI v3 (non-verbatim) & 0.00 pp & +9.76 pp \\
Rev AI v3 (legal) & $-$1.37 pp & +7.48 pp \\
\bottomrule
\end{tabular}%
\end{table}

\textbf{Interpretation}: $\Delta$EID is positive (indicating structural burden on non-fluent speakers) whenever there is a mismatch between ASR optimization and enforced convention. The magnitude ranges from 4--10 pp. When ASR and enforcement align, $\Delta$EID approaches zero -- not because injustice disappears, but because all groups are equally well-served by the matching convention. The injustice lies in the \textit{choice} of which convention to enforce, not in any particular system's performance.

\section{Extended Related Work}\label{sec:extended-related}

\subsection{Fairness Metrics}

Prior work on algorithmic fairness has largely focused on defining constraints over predictions relative to a fixed ground truth (Table~\ref{tab:fairness-metrics}). Metrics such as disparate impact, demographic parity, equalized odds, and equal opportunity formalize different normative commitments -- equalizing acceptance rates, error rates, or true positive rates across protected groups -- while taking the ground truth label $Y$ as given.

This literature has produced valuable insights into trade-offs among fairness criteria and has shaped practice across domains including credit scoring, hiring, criminal justice, recommender systems, and language technologies across large language models, recommender systems, and automatic speech recognition models
\cite{pessach2022review, wang2023survey, dheram2022toward}, and many debiasing methods have been developed to optimize for fairness performance \cite{gao2025sprec, li2024steering, dong2024disclosure, lin2024towards}. However, these metrics share a common structural assumption: reference monism. They presuppose that the labels against which systems are evaluated are objective, determinate, and independent of social context.

Our work departs from this paradigm at a prior level of abstraction. Rather than asking whether predictions are distributed fairly given a ground truth, we ask how the choice of ground truth itself structures downstream fairness assessments. In ASR evaluation, the ``true label'' is not a natural kind but the output of a transcription convention -- one that encodes normative judgments about which features of speech matter and which may be discarded.
This is an instance of a broader problem in fair ML: fairness-relevant constructs must be operationalized via a measurement model, and mismatches between a construct and its operationalization produce downstream harms \cite{jacobs2021measurement, selbst2019fairness}. Prior work has shown that collapsing multiple legitimate annotation perspectives to a single ground truth introduces representational biases \cite{denton2021whose, plank2022problem, davani2022dealing}. Label distribution learning \cite{geng2016label} motivates treating ambiguous labels as distributions rather than point estimates; our setting differs in that label multiplicity arises not from epistemic uncertainty but from the co-existence of multiple normatively legitimate transcription conventions \cite{aroyo2015truth}. 

As a result, conventional group fairness metrics are ill-suited to detect a distinct form of injustice in speech technologies: systematic disparities induced by the enforcement of a single transcription convention when multiple legitimate conventions exist. 
Standard group fairness criteria -- equalized odds, equal opportunity, and demographic parity -- condition on a fixed ground truth and ask whether errors are distributed equitably \cite{hardt2016equality, dwork2012fairness, mehrabi2021survey}. They cannot surface harms that are constituted by the choice of ground truth itself \cite{blodgett2020language}.
Two systems may satisfy equalized odds relative to a clean reference while nonetheless imposing unequal interpretive burdens on speakers whose communicative practices diverge from that convention.

Our approach therefore complements, rather than replaces, existing fairness metrics. The quantities we introduce -- EID and $\Delta$EID -- do not constrain prediction distributions. Instead, they measure the evaluation cost of reference monism: \textbf{the extent to which a group is penalized solely because evaluation restricts the space of legitimate interpretations.} This shifts the locus of fairness analysis from model behavior alone to the institutional practices that define correctness.

In this sense, EID operates upstream of standard fairness metrics. It diagnoses when disparities attributed to model bias are in fact artifacts of evaluative infrastructure. Once plural ground truths are acknowledged, conventional group fairness metrics can be meaningfully reapplied within each interpretive framework. Without this step, fairness assessments risk reintroducing the very interpretive exclusions they aim to measure.

\begin{table*}[!t]
\centering
\caption{Overview of common group fairness metrics \cite{pessach2022review}. $S$ denotes a protected attribute, $\hat{Y}$ the predicted label, $Y$ the true label, and $\varepsilon$ an allowed tolerance.}
\label{tab:fairness-metrics}
\small
\renewcommand{\arraystretch}{1.25}

\begin{tabularx}{\textwidth}{
  >{\raggedright\arraybackslash}p{3.2cm}
  >{\raggedright\arraybackslash}p{3.2cm}
  >{\raggedright\arraybackslash}X
}
\toprule
\textbf{Metric} & \textbf{What it Enforces} & \textbf{Formal Definition} \\
\midrule
\textbf{Disparate Impact}
& Similar \emph{positive prediction rates} across groups
& $\Pr(\hat{Y}=1 \mid S \neq 1)\,/\,\Pr(\hat{Y}=1 \mid S=1) \ge 1 - \varepsilon$
\\
\textbf{Demographic Parity}
& Equality of positive predictions, regardless of ground truth
& $\lvert \Pr(\hat{Y}=1 \mid S=1) - \Pr(\hat{Y}=1 \mid S \neq 1) \rvert \le \varepsilon$
\\
\textbf{Equalized Odds}
& Equal error rates (FPR and TPR) across groups
& $\lvert \Pr(\hat{Y}=1 \mid S=1,Y=y) - \Pr(\hat{Y}=1 \mid S \neq 1,Y=y) \rvert \le \varepsilon,\ \forall y\in\{0,1\}$
\\
\textbf{Equal Opportunity}
& Equal true positive rates across groups
& $\lvert \Pr(\hat{Y}=1 \mid S=1,Y=1) - \Pr(\hat{Y}=1 \mid S \neq 1,Y=1) \rvert \le \varepsilon$
\\
\bottomrule
\end{tabularx}%
\end{table*}

\subsection{Fairness Metrics and the Level of Analysis}
\label{sec:Fairness Metrics and the Level of Analysis}

Let $x \in \mathcal{X}$ denote a speech signal and $g \in \mathcal{G}$ a speaker group.
Let $h : \mathcal{X} \rightarrow \mathcal{Y}$ be an ASR system producing a hypothesis
$h(x)$.
Let $p \in \mathcal{P}$ denote a transcription policy (or convention), and let
$r^{(p)}(x)$ be the reference transcript induced by policy $p$.
Finally, let $\ell: \mathcal{Y} \times \mathcal{Y} \rightarrow \mathbb{R}_{\geq 0}$ be an evaluation loss function mapping a hypothesis--reference pair to a non-negative scalar; in the ASR setting, $\ell$ is instantiated as WER, so $\ell(h(x),r^{(p)}(x)) = \text{WER}(h(x),r^{(p)}(x))$.

\begin{table*}[!t]
\small
\centering
\caption{Evaluation practices differ in what is treated as variable versus fixed. Only plural ground truth varies the interpretive framework itself.}
\label{tab:evaluation-comparison}
\begin{tabular}{p{3.1cm} p{3.3cm} p{3cm} p{4cm}}
\toprule
\textbf{Practice} & \textbf{What varies} & \textbf{What is held fixed} & \textbf{Formal object of analysis} \\
\midrule
Multiple datasets &
Input distribution $x \sim \mathcal{D}_g$ &
Ground truth $Y$, Policy $p^\star$ &
Generalization gaps: $\mathbb{E}_{x \sim \mathcal{D}_g}[\ell(h(x), r^{(p^\star)}(x))]$ \\

Group fairness metrics &
Prediction distribution $\hat{Y} \mid g$ &
Ground truth $Y$, Policy $p^\star$ &
Group performance gaps: $\Pr(\hat{Y}=1 \mid g, Y=y)$ \\

Multiple annotators &
Annotator $a$ &
Ground truth $Y$, Policy $p^\star$ &
Variance in $r^{(p^\star)}_a(x)$ \\

Robustness testing &
$x \rightarrow x'$ or $Y \rightarrow Y'$ &
Policy $p^\star$ &
Stability of $\ell(h(x), Y)$ \\

Plural ground truth &
Policy $p \in \mathcal{P}$ &
Speech event $x \sim \mathcal{D}_g$ &
$\{\ell(h(x), r^{(p)}(x)) : p \in \mathcal{P}\}$ \\
\bottomrule
\end{tabular}
\end{table*}

Standard group fairness metrics differ in both their conditioning structure and their outcome type. Metrics that condition on ground truth -- such as equalized odds, which require $\Pr(\hat{Y} = 1 \mid g, Y = y)$ to be equal across $g$ for both $y \in \{0,1\}$, and equal opportunity, which requires equality only for $y=1$ -- take $Y$ as given and analyze how predictions are distributed relative to it. Demographic parity conditions only on group membership, requiring $\Pr(\hat{Y} = 1 \mid g)$ to be equal across $g$ without reference to $Y$ at all. A further structural difference is that all of these metrics are defined for discrete, typically binary outcomes ($\hat{Y} \in \{0,1\}$) or multi-class classification ($\hat{Y} \in \{1, \ldots,K\}$), where group-conditional probabilities are well-defined. ASR evaluation instead produces continuous-valued losses (WER $\in [0,1]$), for which these probability-based formulations do not directly apply and must be replaced with expectations over group-conditional loss distributions. Our EID and $\Delta$EID quantities are defined directly in terms of continuous WER, bypassing the need to discretize transcription quality. Across all variants -- whether conditioning on $Y$ or not, whether binary or continuous -- the shared assumption our work challenges is that the outcome space itself is fixed: that $Y$, when it appears, refers to a determinate ground truth rather than one of several equally legitimate transcription conventions.

In ASR evaluation, however, the label $Y$ is not primitive. Instead, it is induced by a transcription convention:
\begin{equation}
Y = r^{(p)}(x),
\end{equation}
where different choices of $p \in \mathcal{P}$ encode different, but equally legitimate, interpretations of the same speech event. Conventional fairness metrics therefore presuppose reference monism: the enforcement of a single policy $p^\star$ as the evaluation standard.

Our approach departs from this assumption by allowing $p$ to vary while holding $x$ fixed. Rather than analyzing disparities in $\hat{Y}$ conditional on $Y$, we analyze how evaluation loss itself varies across legitimate interpretive frameworks:
\begin{equation}
\ell(h(x), r^{(p)}(x)) \quad \text{for } p \in \mathcal{P}.
\end{equation}

This shift exposes a form of structural disparity that fairness metrics defined over $(\hat{Y}, Y)$ cannot detect: group-dependent regret induced by the institutional choice
to collapse interpretive plurality into a single enforced convention.
The quantities introduced in this paper (EID and $\Delta$EID) formalize this regret, measuring the evaluation burden imposed on a group solely by the choice of $p^\star$.

In short, group fairness metrics study disparities of the form
\begin{equation}
\text{Disparity}(\hat{Y} \mid Y),
\end{equation}
whereas plural ground truth studies disparities of the form
\begin{equation}
\text{Disparity}(\ell(h(x), r^{(p)}(x)) \mid p).
\end{equation}
These analyses operate at different levels and are therefore complementary rather than interchangeable.

\subsection{Distinguishing WER-Range from Adjacent Approaches}\label{sec:alternative-approaches}

One might propose averaging WER across conventions:
\begin{equation}
    \text{WER-Avg}(h, D, P) := \frac{1}{|P|} \sum_{p \in P} \text{WER}_p(h, D)
\end{equation}
We resist this for several reasons. Averaging implicitly weights all conventions equally, which is itself a normative choice -- why should verbatim and non-verbatim count the same when clinical contexts privilege the former and accessibility applications may prefer the latter? Averaging also obscures the range: two systems with identical WER-Avg but different ranges (e.g., [10\%, 14\%] vs.\ [5\%, 19\%]) have meaningfully different characteristics. Most fundamentally, the problem is not \textit{which} number we report but that we report \textit{a} number as if it were objective. Averaging produces another single number, maintaining the illusion of determinate ground truth. WER-Range preserves the plurality that averaging collapses, shifting the question from ``what is the true accuracy?'' to ``how does performance vary across legitimate interpretive frameworks?''

A different objection asks whether WER-Range is simply cross-dataset evaluation by another name. It is not. Cross-dataset evaluation varies the \textit{input distribution} -- different speakers, recording conditions, demographics -- while holding ground truth fixed. It asks: how well does this system generalize across speech populations? Plural ground truth varies the \textit{interpretive framework} while holding input constant. The same audio, from the same speaker, is evaluated against different conventions. It asks: how much does apparent accuracy depend on which standard defines correctness? These analyses are orthogonal and locate performance variation differently. Cross-dataset variation is typically attributed to the system; convention variation cannot be, since a verbatim-optimized system scoring poorly against non-verbatim references is not malfunctioning but misaligned with the evaluative standard. Table~\ref{tab:evaluation-comparison} formalizes this distinction across five evaluation practices\footnote{We characterize multiple-annotator variation as variance in $r^{(p^\star)}_a(x)$ as a default, following standard inter-rater reliability frameworks \cite{krippendorff2011computing}. However, annotator disagreement can be analyzed in other ways -- e.g., as label distributions \cite{uma2021learning} or as evidence of genuine interpretive ambiguity. Our point is that even under the richest treatment of annotator disagreement, the policy $p^\star$ itself remains fixed; plural ground truth relaxes this constraint.}. This differs from multiple-annotator approaches in a crucial respect: having several annotators each follow the same policy $p^\star$ produces a variance within a single interpretive framework -- useful for measuring inter-rater reliability -- but does not reveal how performance changes when the evaluative standard itself changes. Plural ground truth varies the policy $p$, not just the annotator; it asks not ``how consistently does this convention apply?'' but ``what happens when we evaluate under a different convention altogether?''

\section{Implementation Roadmap}\label{sec:roadmap}

For organizations committed to fair evaluation despite resource constraints, we propose staged implementation:

\textbf{Stage 1: Convention transparency} (minimal cost). Replace ``WER'' with ``WER (under convention $p$)'' in benchmark papers, system cards, and documentation \cite{gebru2021datasheets, papakyriakopoulos2023augmented}. This requires no additional annotation but makes interpretive commitments visible. Concretely: the Open ASR Leaderboard should adopt convention-labeled WER reporting; Datasheets for Datasets \cite{gebru2021datasheets} and speech dataset documentation \cite{papakyriakopoulos2023augmented} should include transcription convention as a required field; and ASR system cards should report performance under multiple conventions when targeting diverse user populations.

\textbf{Stage 2: Diagnostic evaluation} (moderate cost). Produce multi-reference transcripts for a representative subsample (10--20\% of data). If WER-Range is narrow, justify single-reference evaluation; if wide, proceed to Stage 3.

\textbf{Stage 3: Strategic plural ground truth} (higher cost). Produce multiple references for populations where prior evidence suggests convention-dependence: clinical speech, non-standard dialects, child speech, second-language speakers.

\textbf{Stage 4: Infrastructure investment} (long-term). Contribute multi-reference datasets to public repositories; advocate for funding agencies to support annotation infrastructure as essential research infrastructure.

We do not dismiss cost concerns, but cost should be invoked to prioritize resources rather than justify epistemic monism. When evaluation practices systematically disadvantage marginalized speakers, the question is not ``can we afford to fix this?'' but ``who bears the cost of not fixing this?'' -- the speakers whose contributions are rendered unintelligible by frameworks refusing to recognize their legitimacy.

\bibliographystyle{acm}
\bibliography{aaai2026}

\end{document}